\documentclass[10pt,letter]{article}

\input{preface.tex}
\arxivtrue
\usepackage{amsmath}
\usepackage{amssymb}
\usepackage{authblk}
\usepackage[margin=1in]{geometry}

\begin{document}

\title{
  \sysname{}: Conditional Biomedical Abstract Generation
}

\author[1]{
  Justin Sybrandt
  \thanks{\href{mailto:jsybran@clemson.edu}{jsybran@clemson.edu}}
}
\author[1]{
  Ilya Safro
  \thanks{\href{mailto:isafro@clemson.edu}{isafro@clemson.edu}}
}
\affil[1]{
  School of Computing,
  Clemson University
}

\date{}

\maketitle

\begin{abstract}
  Biomedical research papers use significantly different language and jargon
  when compared to typical English text, which reduces the utility of pre-trained
  NLP models in this domain. Meanwhile Medline, a database of biomedical
  abstracts, introduces nearly a million new documents per-year.  Applications
  that could benefit from understanding this wealth of publicly available
  information, such as scientific writing assistants, chat-bots, or
  descriptive hypothesis generation systems, require new domain-centered approaches.
  A conditional language model, one that learns the probability of
  words given some a priori criteria, is a fundamental building block in many
  such applications.  We propose a transformer-based conditional language model
  with a shallow encoder ``condition'' stack, and a deep ``language model''
  stack of multi-headed attention blocks. The condition stack encodes
  metadata used to alter the output probability distribution of the
  language model stack.  We sample this distribution in order to generate biomedical abstracts 
  given only a proposed title, an intended publication year, and a set of
  keywords.  Using typical natural language generation metrics, we
  demonstrate that this proposed approach is more capable of producing non-trivial relevant
  entities within the abstract body than the 1.5B parameter
  GPT-2 language model.  {\bf Reproducability:} All code, data, pre-trained
  models, and experimental parameters are available online: \repo{}

\end{abstract}

\ifarxiv
\else
%
% The code below should be generated by the tool at
% http://dl.acm.org/ccs.cfm
% Please copy and paste the code instead of the example below.
%

%
% End generated code
%

\fi

\section{Introduction}
\label{sec:introduction}

% Lets build this around hypothesis generation

The biomedical sciences are becoming more data driven due to the increased
availability of experimental data and the democratization of machine learning
algorithms. One subfield of biomedical data science, literature-based discovery~\cite{sybrandt2017},
produces algorithms to automatically identify plausible research directions from
the growing body of scientific literature~\cite{bruza2008literature}. While
these systems have seen early successes aiding biomedical
science~\cite{aksenova2019inhibition,bakkar2018artificial}, these techniques
often lack the interpretability necessary to persuade domain scientists to
pursue algorithmically generated leads. While customizable visualizations aid
significantly~\cite{spangler2015accelerating}, many researchers would prefer a
textual description to accompany generated hypotheses. Thinking much further
into the future, if hypothesis generation systems are ever going to function as
automated scientists in their own right, they will require the ability to
generate textual arguments supporting their own ideas.

Today, modern deep-learning language generation models can produce text in a
range of contexts. Building off of the transformer
architecture~\cite{vaswani2017attention}, models like BERT~\cite{devlin2019bert}
and GPT/GPT-2~\cite{radford2018improving,radford2019language} have set a new
standard in a range of natural language benchmarks~\cite{wang2018glue}.
Adaptations of these models, such as SciBert~\cite{beltagy2019scibert} and
BioBert~\cite{lee2019biobert}, have retrained the baseline models for
domain-specific tasks in order to advance the state of domain-specific
benchmarks as well~\cite{hirschman2005overview,peng2019transfer}. However,
these domain-specific models are not designed for natural language generation
(NLG). Models like GPT are trained to generated text in the language of typical
English writing, and we demonstrate below that these generations are ill-suited
for the jargon-filled particular language used by biomedical scientists.

Compounding these challenges around generating biomedical text, much less work
has focused on \textit{conditional} generation, wherein the output language
distribution is affected by a priori knowledge. Older text captioning
systems~\cite{you2016image} follow a similar approach using sequence models
informed by image encodings.  However, more control over generated text is
necessary for applications like hypothesis generation systems, where semantic
information detected by the system should be leveraged in an automatically
produced argument.  Modern systems trained outside the biomedical domain, such
as Ctrl~\cite{keskar2019ctrl} allow for some conditions, but lack the
flexibility needed to capture sets of semantic information. More generalizable
methods, such as those produced by variational
auto-encoders~\cite{hu2017toward}, can capture rich latent language semantics,
but cannot straightforwardly encode domain-based information, such as a set of
keywords one wishes to include in the output text.

A language model that can enable complex domain-specific applications, such as
hypothesis generation, therefore requires a new approach. This technique should
accept an arbitrary set of semantic criteria as a condition, should be aware of
domain-specific entities and jargon, and should produce text that would be
expected by biomedical scientists.

In this paper we propose \sysname{}, a conditional biomedical abstract
generation model that seeks to address the above requirements. This transformer
model includes a shallow encoder stack to encode qualities of the condition, and
an deep decoder stack to produce a high quality language model. We train this
model using semi-supervised multi-task generative pre-training, wherein to
minimize our proposed objective function, the model must predict successive
tokens, parts of speech, dependency tags, as well as entity labels. We train
this model using over 20-million biomedical records provided by the National
Library of Medicine (NLM) through the Medline database. Each record consists of
a title, abstract, publication year, and an optional set of author-provided
keywords. Text processing and annotations are provided by a biomedical NLP model
trained on the ``BIONLP13CG'' BioCreative training
set~\cite{hirschman2005overview}. This pre-trained domain-specific model allows
the \sysname{} model to apply the knowledge gain from the relatively small
human-annotated dataset to the larger set of unstructured text present in
Medline.

We train the proposed model by sampling textual windows 
from within MEDLINE  abstracts.
The publication date, and any author-supplied Medical Subject Headings (MeSH
terms, a set of biomedical keywords and phrases) form the condition.  The sampled window serves as input to the decoder
stack. Windows are split into subword units using the unigram
subword-regularization algorithm~\cite{kudo2018subword}. Using masked-self
attention, we train the model to predict each subword $i+1$ using only the
condition and tokens $1,\dots,i$.

To the best of our knowledge, this work is the first attempt to design a biomedical abstract generator.
Therefore, without a direct point of comparison, we leverage the 1.5-billion parameter ``huge''
version of GPT-2 to compare against. As this language model was trained on a
range of online data sources, such as the BooksCorpus and English Wikipedia, it
is a disadvantage in our domain-specific task. However, the authors find that this
model is capable of a range of specific tasks across domains, such as language
translation, question answering, and commonsense
reasoning~\cite{radford2019language}. Furthermore, other work has even found
that the GPT-2 language model can function as a general purpose knowledge
base~\cite{petroni2019language}. For these reasons, we can expect GPT-2 to be a
relevant, albeit disadvantaged, point of comparison.

When generating an abstract during evaluation, we formulate a human written
title, as well as relevant condition information where applicable, for model
input. We then sample each model's subword probability distribution for each
generated result until the new abstract is written.  We evaluate
computer-generated abstracts based on their ability to produce relevant
$n$-grams that occur in the human-written abstract associated with the input
title. We leverage a range of NLG metrics~\cite{sharma2017nlgeval}, such as
Bleu, METEOR, ROUGE-L and CIDEr, including a version of CIDEr that omits input
$n$-grams from consideration. Through all considered metrics we quantitatively
demonstrate increased performance through the use of \sysname{}.  Qualitatively,
we present full-abstracts, as well as a handful of sentences for assorted
generations, which show the ability of our proposed model to capture the
overarching flow of scientific summaries. We additionally demonstrate the
ability for condition keywords to influence model generations by producing a
varied set of completions for the seed-phrase, ``\textit{In this study, we
found...}''

\noindent{\bf Our contribution:} We present \sysname, a transformer-based
language model for conditional biomedical abstract generation. Trained using
Medline records and informed by semi-supervised domain-specific annotations,
this model captures biomedical jargon, entities, and pattern of scientific
discussion. We compare generated abstracts against the 1.5B
parameter GPT-2 language model, and demonstrate a superior ability to produce
relevant $n$-grams across a range of NLG metrics.

All code, data, pre-trained models, preprocessing pipelines, and experimental
parameters are available online\footnote{\repo{}}. We additionally supply a set
of over 13,000 automatically generated abstracts for a wide range of test-set
titles. Using the generalizable precondition approach presented here, we hope to
enable future applications, such as descriptive hypothesis generation. However,
we are also cognisant of the potential for abuse surrounding high quality
domain-specific language models. We discuss these concerns further in
Section~\ref{sec:future_work}.

\section{Background}
\label{sec:background}

While recent {\bf language models} receive a newfound popularity in proportion
to their surprising capacity across a range of tasks~\cite{radford2019language},
their study predates modern machine learning techniques~\cite{bengio2003neural}.
Formally, a language model is a probabilistic model that captures the conditional
probability of each next element in a sequence given all prior elements.
Specifically, this is described by the function:
\begin{equation}
  \prob(s) = \prod_{i=1}^{n} \prob(s_i | s_1,\dots,s_{i-1})
  \label{eq:language_model}
\end{equation}

Here, $s$ is a sequence of $n$ elements. The probability of observing sequence
$s$ is determined by the product of the conditional probabilities of observing
each token $s_i$ given all prior tokens. These models can generate new text by
iteratively sampling new elements from the probability distribution
$\prob(s_{i+1}|s_1,\dots,s_i)$.

The conditional language model introduces a new term $c$ into the above
equation. The condition can allow applications to alter the resulting sequence
based on a priori knowledge~\cite{hu2017toward}. 
Formally, the conditional language model is defined as:
\begin{equation}
  \prob(s|c) = \prod_{i=1}^{n} \prob(s_i | s_1,\dots,s_{i-1},c)
  \label{eq:condition_language_model}
\end{equation}

Modern neural network language models~\cite{radford2019language,keskar2019ctrl},
model these probability distributions by minimizing the negative log-likelihood
of these distributions over a large training set of sequences:
\begin{equation}
\begin{split}
  \mathcal{L}\left(
    \left(s^{(1)}, c^{(1)}\right),
    \dots,
    \left(s^{(m)},c^{(m)}\right)
  \right) \\=
    -\sum_{j=1}^{m}
      \sum_{i=1}^{n}
        \log \prob_\theta \left(
          s^{(j)}_i | s^{(j)}_1,\dots,s^{(j)}_{i-1}, c^{(j)}
        \right)
\end{split}
\end{equation}

Here, $\prob_\theta$ indicates the parameterized model that approximates the
language model distribution. Modern systems often use the transformer
architecture~\cite{radford2019language,keskar2019ctrl,wang2019bert} for
state-of-the-art quality estimating $\prob_\theta$.

\noindent{\bf The transformer}~\cite{vaswani2017attention},
a sequence-to-sequence model built through multi-headed attention layers,
has been customized for a number of NLP tasks, as best demonstrated by
BERT~\cite{devlin2019bert}, GPT-2~\cite{radford2019language}, and a range of
notable follow-ups~\cite{raffel2019t5,sun2019ernie,liu2019roberta}.
Conceptually, the attention mechanism works by learning
multiple weighted averages per-element of the input sequence.
Specifically, this includes three projections of each
element's embedding, represented as packed matrices: $Q$, $K$, and
$V$.  Each projection functions differently, with $Q$ acting as a ``query''
that is compared against ``keys'' $K$ and ``values'' $V$. The specific mechanism
is defined as follows, with $d_k$ representing the dimensionality of each $Q$
and $K$ embedding:

\begin{equation}
  \text{Attention}(Q, K, V) =
  \text{softmax}\left(\frac{QK^\intercal}{\sqrt{d_k}}\right)V
\end{equation}

The ``multi-headed'' aspect of the transformer indicates that the self-attention
mechanism is applied multiple times per-layer, per-element of the sequence.
These multiple heads are then recombined through a feed-forward layer:

\begin{equation}
\begin{aligned}
  \text{MultiHead}(X, Y) &= [h_1;\dots;h_k]W^{(4)} \\
  \text{where } h_i &=
  \text{Attention}\left(XW_i^{(1)},YW_i^{(2)},YW_i^{(3)}\right)
\end{aligned}
\end{equation}

The transformer model presented by Vaswani et al.~\cite{vaswani2017attention}
use the attention attention
mechanism in three different ways. Within the encoder stack, which processes
the input sequence in their proposed sequence-to-sequence model, the
$K$, $Q$, and $V$ embeddings all come from the same sequence of tokens. This is
referred to all ``self attention.'' In the decoder stack, the part of the model
that uses the encoder output to generate a new sequence, these embedding
matrices are masked during the attention function such that the output embedding
for position $i$ can only depend on prior elements. This is called ``masked self
attention''. Following this operation, each decoder embedding is attended with all
of the encoder embeddings. Specifically, $Q$ values are derived from the
decoder, while $K$ and $V$ values depend on the encoder. We refer to this
operation as ``Encoder-Decoder Attention.'' Note that BERT~\cite{wang2019bert}
uses only the encoder self-attention layers, while
GPT-2~\cite{radford2019language} uses the decoder's masked self-attention
layers. The work presented here uses all three.

The multi-head components are combined with a feed-forward operation, denoted FF, that
projects the concatenated embedding into a larger dimensionality, applies the
ReLU activation function, and then reduces back to the set embedding rank:
\begin{equation}
  \text{FF}(X) = \text{max}(0, XW)W'
\end{equation}

Then, combined with a learned layer-wise normalization, these components combine
to form encoder and decoder blocks. Omitting the standard dropout between each operation, the encoder block is defined as:
\begin{equation}
\begin{aligned}
  \mathcal{E}(X) &= \text{LayerNorm}(\text{FF}(\alpha) + \alpha) \\
  \alpha &= \text{LayerNorm}(\text{MultiHead}(X, X) + X)
\end{aligned}
\end{equation}

while the decoder block is defined as:
\begin{equation}
\begin{aligned}
  \mathcal{D}(X, Y) &= \text{LayerNorm}(\text{FF}(\alpha) + \alpha) \\
  \alpha &= \text{LayerNorm}(\text{MultiHead}(\beta, Y) + \beta) \\
  \beta &= \text{LayerNorm}(\text{MultiHead}(X, X) + X)
\end{aligned}
\end{equation}

\ifthesis
%tokenization
\noindent{\bf Tokenization} chunks an input sequence of characters into input
for a transformer-based model.  BERT leverages the WordPiece
algorithm~\cite{wu2016google}, which first learns to identify a predetermined
number of character-groups from a sample of text in order to minimize the
expected number of character groups per sentence. The fact that practitioners
can tune the number of tokens in a WordPiece tokenization of critical for
lowering the overall vocabulary words, and ultimately the size of the model.
This approach also allows the model to more easily adapt to out-of-vocabulary
words, as infrequent words can simply be constructed by assembling smaller
word-chunks (often the chunks containing a single
character)~\cite{sennrich2015neural}. While the WordPiece algorithm itself is
proprietary, SentencePiece is an official open-source implementation.

Many groups have worked to endow transformer-based language models with
domain-based information. In the field of scientific language, two major models
have been proposed: SciBERT from AllenNLP~\cite{beltagy2019scibert}, and
BioBERT from Korea University in Seoul~\cite{lee2019biobert}. SciBERT is trained
on over one-million papers from SemanticScholar.org, and constructed to
completed named entity recognition, PICO Extraction, Text Classification,
Relation Classification, and Dependency Parsing. For each of these tasks,
training data is provided by relatively small human annotated datasets. Improved
performance comes from initial pretraining done on the base of the model, in the
same manner as was performed for the original BERT. From there, the base model
can be used to instantiate fine-tuned version of SciBERT, each with different
``task-heads,'' which learn to associate the fundamental semantic content of the
base SciBERT model with the particular task at hand. BioBERT performs a similar
procedure, focusing on texts available from Medline and PubMedCentral, as well
as English Wikipedia and the Books Corpus. Then, after being pretrained on all
four datasets, BioBERT fine-tunes for named entity recognition, relation
extraction, and question answering. Again, the datasets used for fine-tuning are
significantly smaller than the datasets used for the BioBERT pretraining phase.
In both cases, SciBERT and BioBERT demonstrate superior performance in their
respective tasks.
\fi

\section{Multi-Conditional Language Model}
\label{sec:methods}

The \sysname{} model follows the transformer
architecture~\cite{vaswani2017attention} with a shallow
``condition'' encoder, and a deep ``language model'' decoder. This model is
depicted in Figure~\ref{fig:model}. The condition is
specified as a set of embeddings that enable a high degree of control. To
capture information that is particular to language within biomedical domain, we
add terms in our objective representing not only elements of the textual
sequence, but also the part-of-speech, dependency tags, and entity class labels
associated with each textual element. For each class of prediction, we minimize
the sum of negative log likelihood:

\begin{figure}
    \centering
\ifarxiv
    \includegraphics[width=0.4\linewidth]{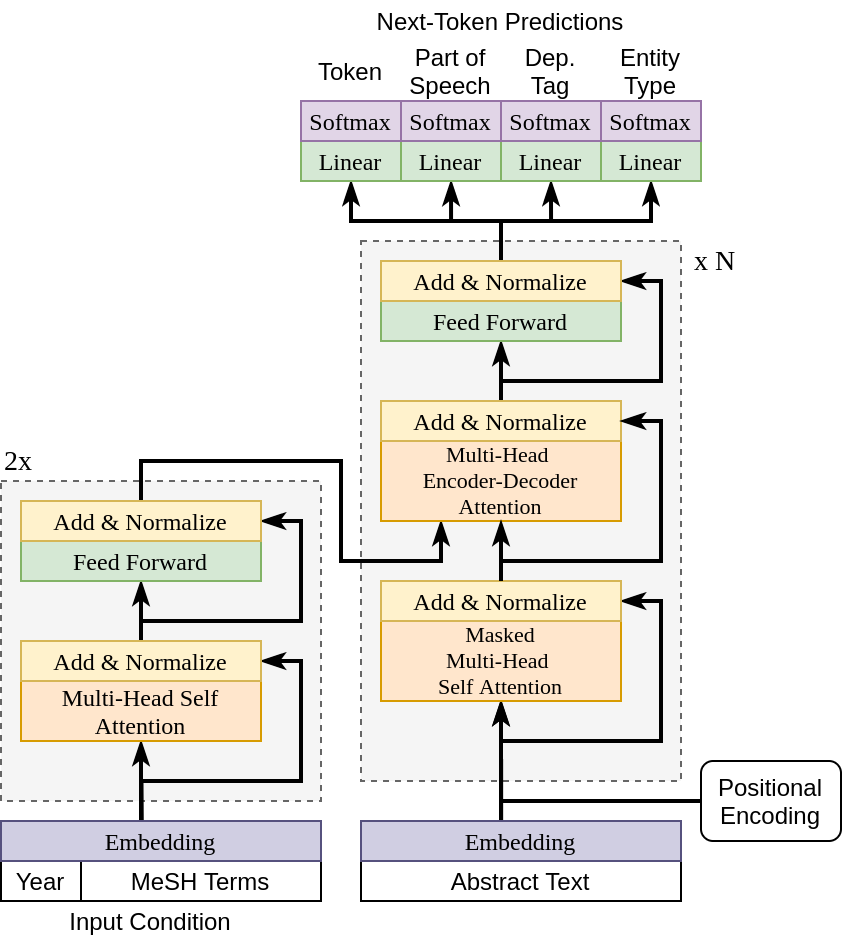}
\else
    \includegraphics[width=0.9\linewidth]{images/model.png}
\fi
    \caption{Abstract Generator Model.}
    \label{fig:model}
\end{figure}

\begin{equation}
  \mathcal{L}(t, p, d, e, c) =
  L_T(t, t, c) + L_P(p, t, c) + L_D(d, t, c) + L_E(e, t, c)
  \label{eq:overall_loss}
\end{equation}

where $t = t_1,\dots,t_n$ are the set of ground-truth textual elements,  each with
associated $p_i\in p $ part-of-speech tags, $d_i\in d$ dependency labels,
$e_i\in e$ entity labels. The term $c=c_1,\dots,c_m$ indicates the set of
conditions associated with $t$, and captures information such as metadata
keywords and the publication year of the ground truth elements. Each term of
(\ref{eq:overall_loss}) follows the form of:

\begin{equation}
\begin{aligned}
  L_{[\cdot]}(\ell, t, c) &=
    \sum_{i=1}^{n}
      -p^{(i)}_{\ell_i}
      + \log\left(
        \sum_{j\neq i} \exp\left(p^{(i)}_j\right)
      \right)\\
  \text{where }p^{(i)} &= \text{softmax}\left(
    \mathcal{H}\left(
      \{t_1,\dots,t_{i-1}\},
      c
    \right)
    W_{[\cdot]}
  \right)
\end{aligned}
\end{equation}

where the symbol $[\cdot]$ is replaced by $T$, $P$, $D$, or $E$ for each
classification
objective.
The sequence $\ell$ indicate the ground-truth labels associated with each
element of $t$ with respect to the particular classification task.
Additionally, $\mathcal{H}(t, c)$ is the proposed transformer model,
which accepts all text elements $\{t_1,\dots,t_{i-1}\}$ and $c$ in order to produce an encoding
for $t_i$. This model is defined as:

\begin{equation}
\begin{aligned}
  \mathcal{H}(t,c) &= D_d \\
  D_{i+1} &= \mathcal{D}(D_i, E_e) \text{ and } D_0 = t + \text{PE}\\
  E_{i+1} &= \mathcal{E}(E_i) \text{ and } E_0 = c \\
\end{aligned}
\end{equation}

Here, PE references the positional encoding defined by the sinusoidal function
presented in~\cite{vaswani2017attention}. Each input element of $t$ and $c$ is
first assigned an input encoding and put through their respective stacks of
encoder and decoder layers. Input encodings are provided by an embedding table
that begins randomly initialized. We determine textual elements through the
unigram word-part tokenizer~\cite{kudo2018subword}, and contextual elements
consist of a learned embedding per-publication year, as well as embeddings for
each Medical Subject Heading (MeSH term). These input factors are described in
father detail in Section~\ref{sec:data_prep}.

\noindent{\bf Hyperparameters.} We selected hyperparameters similar to the GPT-2
``medium'' model. This includes an embedding dimensionality of $d_k=1,024$,
$k=16$ attention heads per multi-headed attention layer, $e=2$ encoder blocks,
$d=16$ decoder blocks, a fully-connected size of 3,072, and an inner-block
dropout rate of 0.1. We additionally use a max sequence length of $n=128$. Our
set of initial embeddings contains 16,000 text tokens, 48,133 MeSH headings, and
230 year embeddings.

\noindent{\bf Optimization.} We minimize $\mathcal{L}$ using the large-batch
optimizer LAMB~\cite{you2019large} across 40 Nvidia V100 GPUs using an effective
batch size of 480. We selected a learning rate of 0.001, with a 500-batch linear
warm up. We check pointed the model each epoch after viewing 5\% of the training
data (about 700,000 abstracts). Note that each time an abstract is viewed, we
select from it a different training window. We trained this model for 72 hours
using PyTorch Lightning~\cite{falcon2019pytorchlightning} to aid in the
distribution and check pointing.

\section{Data Preparation}
\label{sec:data_prep}

\begin{figure*}
\centering
\begin{subfigure}{.5\textwidth}
    \centering
    \includegraphics[width=0.9\linewidth]{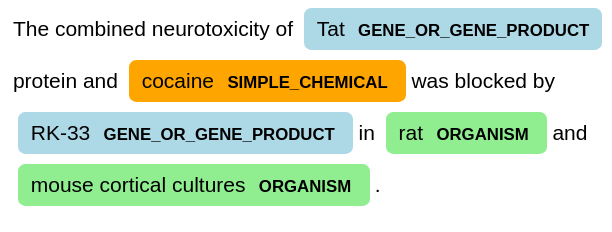}
    \caption{Typed entity recognition.}
\end{subfigure}%
\begin{subfigure}{.5\textwidth}
    \centering
    \includegraphics[width=0.9\linewidth]{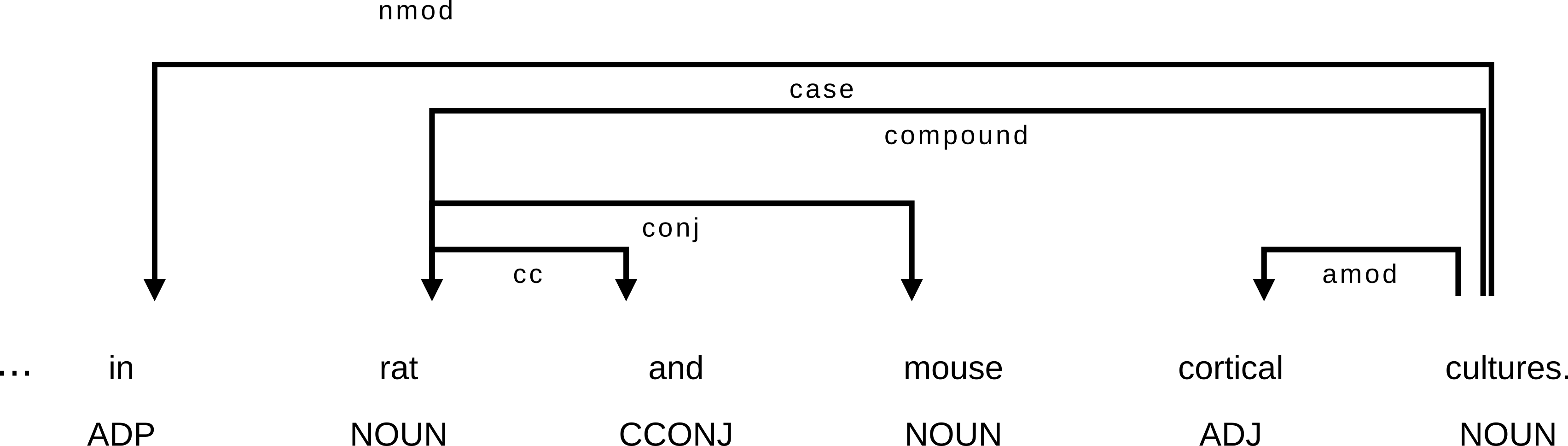}
    \caption{Dependency tags and parts of speech.}
\end{subfigure}
\caption{
  \label{fig:scispacy_annotations}
  Annotations provided by ScispaCy ``BIONLP13CG.''
}
\end{figure*}

In order to train the model described above, we collect training samples $(t,c)$
from the set of publicly available biomedical abstracts provided in the MEDLINE
database. This dataset contains publication dates, author-supplied MeSH
terms, titles, and abstracts for mote than 30-million citations. We filter 
for documents that were originally published in English, as well as documents
that contain at least one non-title sentence. Documents without metadata
keywords are allowed. We split the remaining 20-million abstracts into a
training and test set following a 70-30 split.

Within the domain of biomedical text
mining, there are relatively few annotated training
sources~\cite{hirschman2005overview,peng2019transfer}. To endow the \sysname{}
model with biomedical-domain knowledge, we annotate the entire MEDLINE training
set using an NLP model trained on a smaller annotated training set. Because we
leverage patterns mined from a small human-annotated dataset to gain broader
insights across a vast unstructured dataset, we refer to our overall approach as
semi-supervised. The ScispaCy model~\cite{neumann2019scispacy} trained on the
``BIONLP13CG'' BioCreative dataset~\cite{hirschman2005overview} provides our
biomedical NLP model. This model was selected because it produces the widest
range of entity labels when performing named entity recognition, which consist
of: cancer, organ, tissue, organism, cell, amino acid, gene or gene product,
simple chemical, anatomical system, immaterial anatomical entity, multi-tissue
structure, developing anatomical structure, organism subdivision, and cellular
component. We add a class corresponding to ``not an entity'' as well.

Using the ScispaCy model and a cluster of 100 machines, we quickly identify
every token, part-of-speech, dependency tag, and entity label for all 14-million
training-set MEDLINE documents. We depict examples of these automatic
annotations in Figure~\ref{fig:scispacy_annotations}. However, in order to
formulate these textual features for input into the \sysname{} model, we also
leverage the unigram subword regularization method from Kudo et
al.~\cite{kudo2018subword}. This method learns an efficient tokenization
sentences. Each token corresponds to a ``chunk'' of characters, many of which
correspond to subword components. The unigram approach adds a normalization
factor wherein the specific tokenization for each word is probabilistic
determined from the set of ambiguous subword sequences. These subword sequences,
along with special ``start of abstract'' and ``end of abstract''
tokens, create input $t$.

We train the unigram tokenization method on one-million randomly sampled
sentences from the training set, specifying a fixed-size vocabulary of 16,000
subword tokens. We additionally lowercase the entire training corpus, and
enforce that every character within the sampled training set receive its own
token. Using the resulting model, we tokenize the entire training set, and cross
reference the subwords
with the multi-task labels provided by
ScispaCy. This way, each subword token $t_i$ in the training set is associated
with a part-of-speech $p_i$, dependency tag $d_i$, and entity label $e_i$.

Next we index each training-set publication years and author-supplied MeSH
keywords, which form the condition $c$. For publication years, we simply identify
the earliest year within the training set, 1790, and add an index for each year
between then and 2020. We identify over 4-million author-supplied keywords
within MEDLINE, which is prohibitively large for our model to capture. We prune
any keyword that occurs fewer than ten times, reducing that set to a manageable
48,133. We add each to our excising embedding index, which contains nearly
50,000 total embeddings.

\begin{figure}
    \centering
\ifarxiv
    \includegraphics[width=0.7\linewidth]{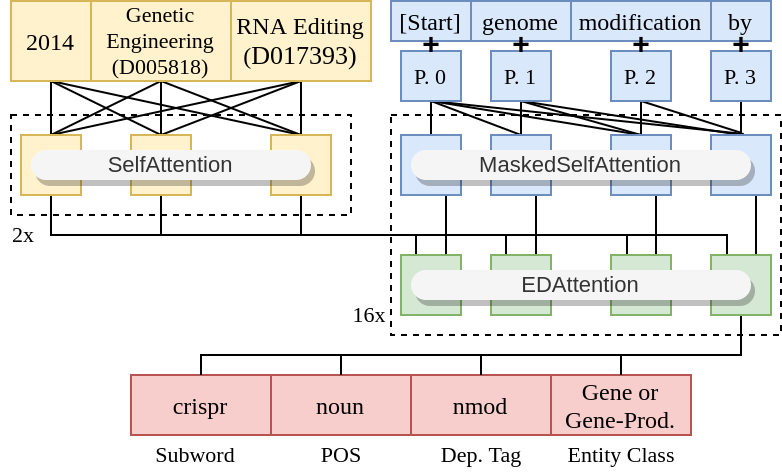}
\else
    \includegraphics[width=0.9\linewidth]{images/example_input.png}
\fi
    \caption{Abstract Generator Example Input.}
    \label{fig:example_input}
\end{figure}

When training, we select a batch of abstracts, and for each abstract we select a
window of $128$ subword tokens to form $t$, restricted such that the first token
of each window corresponds to the first token of a sentence. In addition, we
supply the condition indices $c$. The sequence of labels $\ell$ is formulated by
shifting the subword token window by one token, such that $t_{i-1}$ is used to
predict $t_i,p_i,d_i,$ and $e_i$. An example of model input and output is
depicted in Figure~\ref{fig:example_input}.

\section{Results}
\label{sec:results}

While NLP benchmarks such as GLUE~\cite{wang2018glue} and its biomedical
counterpart BLUE~\cite{peng2019transfer} help researchers compare performance
across a range tasks, we are unaware of a benchmark for the generation of
biomedical abstracts. In lieu of such a dataset, we leverage our held-out
test-set of Medline abstracts, and a set of traditional NLG
metrics~\cite{sharma2017nlgeval}. We generate abstracts by providing a title $t$
and condition $c$ from a test-set abstract. We extend $t$ by sampling from the
resulting probability distribution over subword tokens $p^{(i)}$ until observing
the ``end of abstract'' special token. The quality of the resulting abstract is
quantified for each metric, Bleu~\cite{papineni2002bleu},
METEOR~\cite{lavie2007meteor}, ROUGE-L~\cite{lin2006information}, and
CIDEr~\cite{vedantam2015cider}, by comparing each 
generated
sentence against the set of ``reference'' sentences comprising the corresponding
human-written abstract.

To add context to our reported performance numbers, we also generate text using
OpenAI's recently released 1.5-billion parameter ``huge'' GPT-2
model~\cite{radford2019language}. This model has been shown to excel on a number
of tasks without modification, inducing as a replacement to traditional knowledge
bases~\cite{petroni2019language}. However, as this model was trained to generate
language found online, such as in the BooksCorpus and English Wikipedia, it is
at a disadvantage when generating domain-specific text. Because GPT-2 does not
produce any ``end of document'' indicator, we generate the same number of
subword tokens as present in the human-written counterpart, and truncate the
potential partial sentence at the end of the abstract.

\begin{table*}
  \begin{tabular}{p{16cm}}
    {\bf
      (PMID: 28026780)
      Hierarchically Micro- and Nanopatterned Topographical Cues for Modulation
      of Cellular Structure and Function.
    }\\
    %\toprule
    %{\bf Human} \\
    %\input{figures/abstract_samples/escaped/full_original_1.txt} \\
    \hline
    {\bf\sysname{}} \\
    the ability to integrate multiple physiological cues and thereby mediate many cellular functions is critical for many complex life history processes. despite recent advances in high-throughput imaging of biomolecules and their spatiotemporal integration into dynamic structure and function, the precise structural organization and temporal structure of tissue architecture remains poorly understood. here, we present an efficient system for temporally and spatially mapping micro- and nanopatterned topographical cues in organ-specific spatial and temporal properties using multiple imaging modalities. the micro-/nanopatterned geometrical cues can be localized to cell membranes, cells, and proteins. the spatial and temporal dynamics of these local signals are precisely represented by the cross-correlation function, which forms the basis of a geometrical model that accurately provides spatiotemporal information about the spatial location and spatial coordinate of the labels and their functionalities. the model is also capable of correlating the properties of neural cells within their network without affecting the spatial and temporal organization of their spatial features, as well as those of their surrounding tissue. as an example of this model, cell types grown as multilayers are described.
 \\
    \hline
    {\bf GPT-2 Huge} \\
    [\textbackslash{}n] [\textbackslash{}n] The first example of what happens when an artificial neural network'see' complex features on a computer screen is provided by the vision of people with sight loss. The most recent results (e.g., see Krueger's blog) suggest that a person who has a damaged retinal pathway for visual light-scattering called LID is able to'see' objects using low power laser sources. The finding is unexpected because it suggests an adaptive mechanism for recovering limited visual function. The finding has profound implications for understanding how the retina and brain develop, how vision is learned, and how people can potentially regain impaired vision with new retinal implants. [\textbackslash{}n] The retinotopic map shows which parts of the retina belong to what parts of the visual field (more details in the video below). [\textbackslash{}n] The following images are from a retinotopic map showing a person who has lost some peripheral vision due to an injury in his optic nerve and retinal pigment epithelium. Note that he can make out details when looking toward the left, and when looking toward the right, details become distorted or are obscured.
 \\

    \bottomrule
  \end{tabular}
  \caption{
    \label{tab:full_abstracts}
    Full abstracts generated with respect to the same title.
  }
\end{table*}

We present a full abstract from both \sysname{} and GPT-2 in
Table~\ref{tab:full_abstracts}. Note, newline characters produced by GPT-2 are
replaced with ``[\textbackslash{}n]'' due to space limitations.  In this
example, we observe that the \sysname{} model recovers a set of relevant
biomedical entities.  Unsurprisingly, the model parrots some entities that
appear in the title, such as, ``micro- and nanopatterned topographical cues,''
as well as ``cellular functions'' in this example. However, it is also able to
produce more advanced concepts including ``multiple imaging modalities,'' and
``multiscale substrates'' that do not appear in the title but do appear in the
corresponding human-written abstract (not reproduced here for space concerns,
but is publicly available). The GPT-2 model does recover some biomedical
entities, such as ``damaged retinal pathway'' and ``retinal pigment
epithelium,'' however these keywords are unrelated to the considered document. Other
out-of-context entities such as ``artificial neural network,''  ``computer
screen,'' and reference to a blog reduce the ability of a human reader to extract any meaningful
biomedical information from this text. We find that these example abstracts help
motivate the need for domain-specific language models.

\begin{table*}
  \begin{tabular}{p{3cm}p{12cm}}
    \toprule
    Condition & Response \\
    \midrule
    D003270: Contraceptive Agents
    & ...that, during a prospective observational period, the patients were aware of the possibility of adverse cardiac events.
    \\
    D003634: DDT
    & ...that the aromatic (g)-tse, which is often produced in fruit, is potentially useful to suppress green algae as well as pesticide toxicity.
    \\
    D004042: Unsaturated Dietary Fats
    & ...that vitamin e levels are associated with early childhood health consequences.
    \\
    D006046: Gold
    & ...that the nanoparticles provide improved sensitivity to gold nanoparticles, and they are sensitive to ag-b interaction rather than ca-a interaction.
    \\
    D005395: Fish Oils
    & ...that the combination of pinkland and fish oil intakes (ca-like and ca-like) improves the antioxidant effect of yinneria (tricapsa vul) and that can significantly decrease food intake.
    \\
    \bottomrule
  \end{tabular}
  \caption{
    \label{tab:conditional_generation_table}
    Differing generations of the same prompt given various MeSH preconditions.
    We record the first sentence completing the prompt \textit{``In this study, we
    found...''}
  }
\end{table*}

Because \sysname{} is a conditional language model, we explore the range of
responses the model can produce given different conditions. In
Table~\ref{tab:conditional_generation_table} we present the first sentence
produced by the model for the input ``\emph{In this study, we found...}'' given different
conditions. The results indicate that the condition has a significant impact
in the resulting text. When conditioned with the MeSH term for contraceptive
agents, the model discusses a patient study on cardiac side-effects. The output
conditioned on the pesticide DDT describes fruit and toxicity. The output on
gold describes describes gold-nanoparticle sensitivity. These results
demonstrate the ability for the \sysname{} model to learn domain-specific
research content provided by various keyword preconditions.

\begin{table*}
  \begin{tabular}{p{8cm}p{8cm}}
    \multicolumn{2}{p{16cm}}{\bf
      (PMID: 28029317)
      Laparoscopy to Predict the Result of Primary Cytoreductive Surgery in
      Patients With Advanced Ovarian Cancer: A Randomized Controlled Trial.
    }\\
    \hline
    laparoscopic surgery is the standard treatment for patients with advanced ovarian cancer; however, these patients do not receive a standard palliative regimen.

    &
    J Natl Cancer Inst 2008;100:1567–1572. 24. The focus of this review is the effect of apoE4 levels on the risk of poor surgical outcome in patients with advanced ovarian cancer.

    \\

    &\\
    \multicolumn{2}{p{16cm}}{\bf
      (PMID: 27993387)
      Low vitamin D does not predict statin associated muscle symptoms but is
      associated with transient increases in muscle damage and pain.
    }\\
    \hline
    in clinical practice, patients with moderate-to-severe hypervitaminosis d present with debilitating side effects related to statin use.

    &
    ow vitamin d does not predict statin associated muscle symptoms but is associated with transient increases in muscle damage and pain.

    \\

    &\\
    \multicolumn{2}{p{16cm}}{\bf
      (PMID: 28012718)
      Skin-Resident Effector Memory CD8$+$CD284$-$ T Cells Exhibit a Profibrotic
      Phenotype in Patients with Systemic Sclerosis.
    }\\
    \hline
    systemic sclerosis (ssc) is an inflammatory disease characterized by the infiltration of t cells into skin and skin surfaces. the presence of autoantibodies can lead to the development of cutaneous t-cell hyperactivity.

    &
     J. Clin. Invest. 117 : 2748-2759; Dilating collagen in chronic neuropathic pain. Arch. Neurol. 63 : 983-989

    \\

    &\\
    \multicolumn{2}{p{16cm}}{\bf
      (PMID: 27999935)
      Laparoscopic sentinel node navigation surgery for early gastric cancer: a
      prospective multicenter trial.
    }\\
    \hline
    to compare the feasibility and safety of laparoscopic sentinel node navigation surgery with that of conventional in-field navigation (oif) surgery in the treatment of early gastric cancer (egc).

    &
    Patel S et al. (2003) Age associated factors associated with false-positive result of prognostic biomarkers in prostate and breast cancer.

    \\

    \bottomrule
  \end{tabular}
  \caption{
    \label{tab:abgen_vs_gpt}
    \sysname{} (left) compared to GPT-2 ``huge'' with 1.5B parameters (right).
    Both systems are given the same title as a prompt. \sysname{} receives metadata. Results truncated for space.
  }
\end{table*}

To provide further qualitative comparison between the considered models, we
additionally provide a few first-sentences produced given various test-set
titles in Table~\ref{tab:abgen_vs_gpt}. In these sentences, and across the test set, we observe that \sysname{} produces a
number of scientific cliqu\'{e}s. Most clearly, the model captures biomedical
turns of phrase such as ``in clinical practice.'' Additionally we observe that
it is common for \sysname{} to produce an entity followed by an abbreviation
that it will repeat throughout the text. However, we observe that some
abbreviations are not sensible from a human perspective, such as ``in-field
navigation (oif).'' In these cases, the incorrect abbreviation will still be
repeated by the model. 

Not seen in these first-sentences is a trend for the
model to follow major abstract claims with a fictional $p$-value or sample-size.
We find $p$-values in approximately 10\% of abstracts, with a median value of
$0.02$, and when plotting this distribution of generated $p$-values we find it
matches the expected (and troubling) trend of $p$-values in real-world
science~\cite{head2015extent}.

\begin{figure*}
\centering
\begin{subfigure}{.29\textwidth}
    \includegraphics[width=0.9\linewidth]{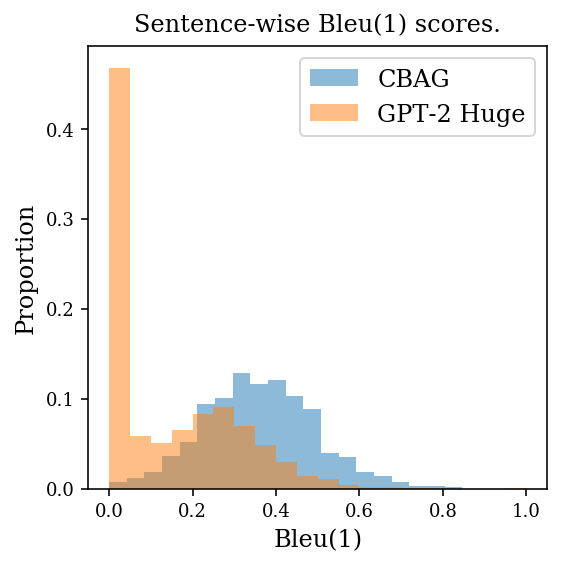}
\end{subfigure}%
\begin{subfigure}{.29\textwidth}
    \includegraphics[width=0.9\linewidth]{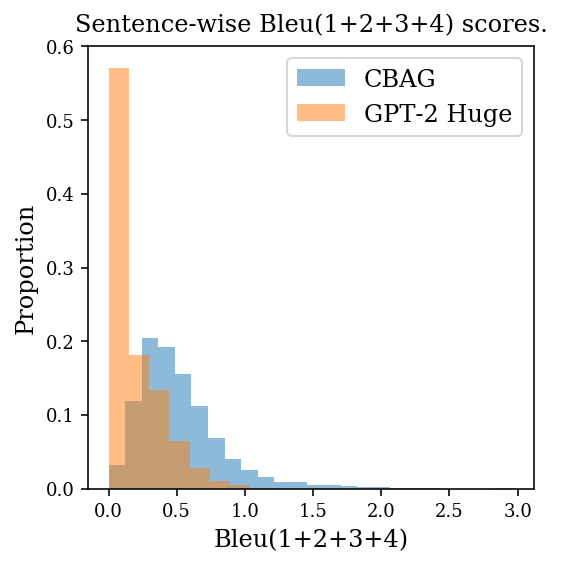}
\end{subfigure}%
\begin{subfigure}{.29\textwidth}
    \includegraphics[width=0.9\linewidth]{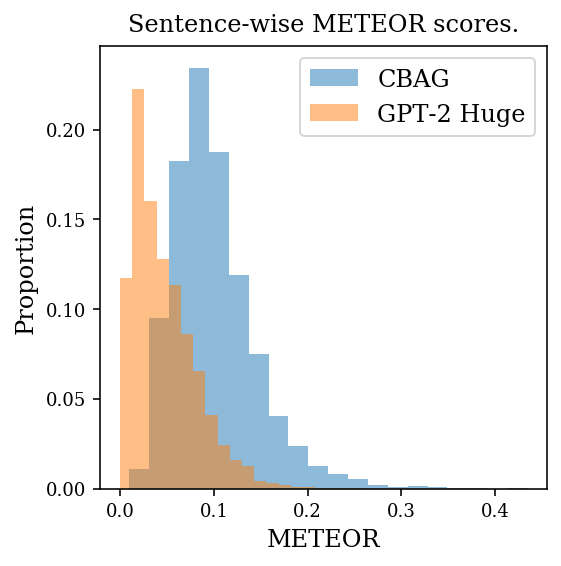}
\end{subfigure}
\begin{subfigure}{.29\textwidth}
    \includegraphics[width=0.9\linewidth]{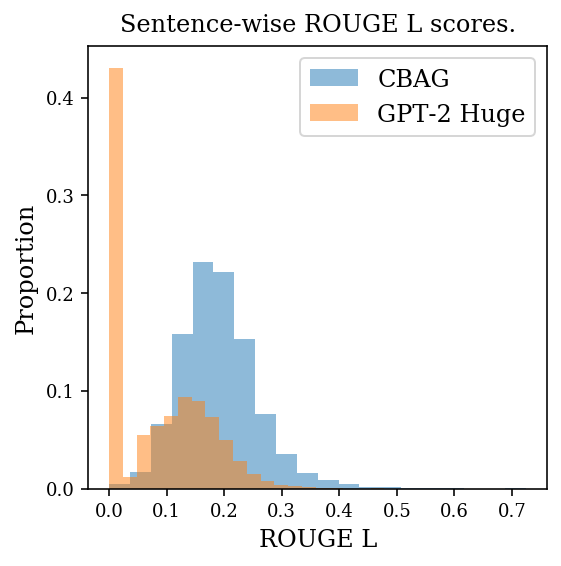}
\end{subfigure}%
\begin{subfigure}{.29\textwidth}
    \includegraphics[width=0.9\linewidth]{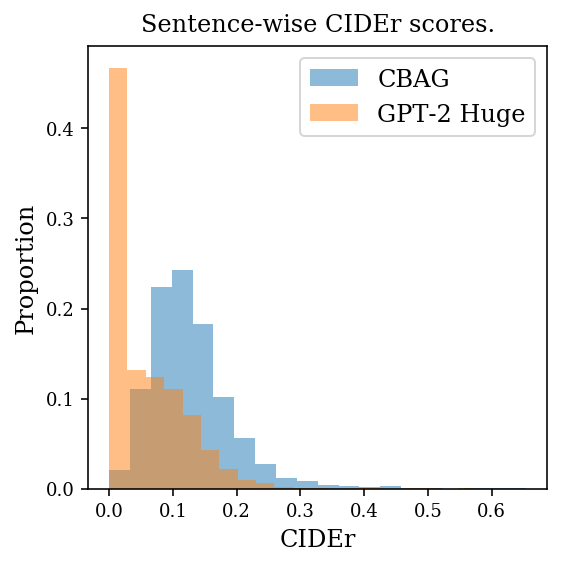}
\end{subfigure}%
\begin{subfigure}{.29\textwidth}
    \includegraphics[width=0.9\linewidth]{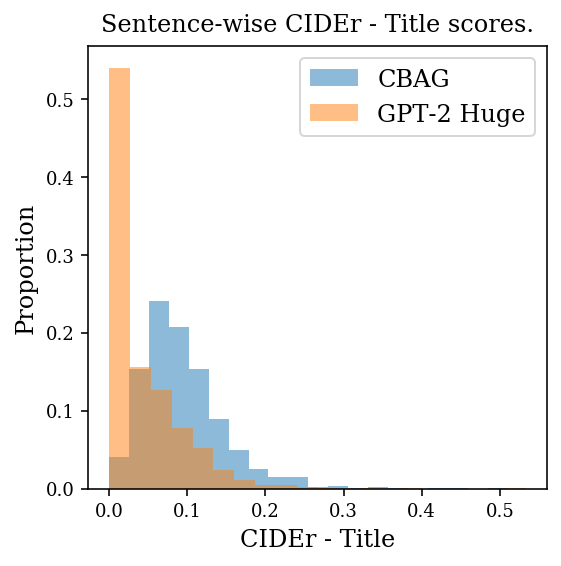}
\end{subfigure}
    \caption{
      \label{fig:score_distribution_table}
      Score distributions per-sentence comparing GPT-2 Huge with \sysname{}.
    }
\end{figure*}

To provide a more rigorous and scalable analysis of \sysname{} generations, we
turn to a collect of NLP metrics, mentioned above. We use two version of Bleu, one
that includes only 1-grams, and one that sums Bleu scores for 1-through-4-grams.
We do not apply smoothing or any additional normalization to Bleu scores in an
effort to reduce unnecessary hyperparameters.  Furthermore, we present two
versions of CIDEr. While both use a sub-sample of training-set abstracts to
approximate $n$-gram document frequency, we also want to determine whether the
generated text can produce uncommon $n$-grams that were not supplied in the
title.
Our ``CIDER-Title'' metric sets the weight of any $n$-gram that appeared in the
title to zero.  The sentence-wise score distribution for all metrics for a
sample of test-set abstracts are depicted in
Figure~\ref{fig:score_distribution_table}, including both scores for \sysname{}
and GPT-2 generations. Note, these histograms are scaled such that all bars for
a particular model sum to one.

We observe that about half of the sentences produced by GPT-2 contain very
little content. As seen in Table~\ref{tab:abgen_vs_gpt}, we see many of
these sentences appear to be in the style of citations, including page numbers
and titles. Therefore, sentences such as ``J Natl Cancer Inst
2008;100:1567–1572. 24.'' are unlikely to recall many relevant $n$-grams. Other
examples, such as the full GPT-generated abstract shown above, seem to discuss
scientific findings from the perspective of an online news outlet covering the
new research.  While the \sysname{} generations are imperfect, they do
score higher, on average, across all considered metrics. In the case of ROUGE-L,
which measures the ability for generated sentences to recall long sub sequences
of text, that many biomedical cliqu\'{e}s are likely easy for \sysname{} to
predict, such as ``the study examined the'' or ``we conclude that the.'' Our
higher METEOR scores, which indicates the ability to recall $n$-grams in the same
order as found in a reference sentence, are also effected by these common
sequences. However, the ``CIDEr-Title'' metric explicitly decreases the weight
of these common $n$-grams, while only considering text that could not be
identified trivially. Our improved performance in this measure, when seen in the
context of our overall improvement, demonstrates the ability for \sysname{} to
produce more relevant and nontrivial biomedical text than the baseline.

\section{Related Work}
\label{sec:related_work}

\ifthesis
\noindent{\bf BERT}~\cite{devlin2019bert} is a transformer-based
model that consists a stack of
unmasked multi-headed self-attention, which means that every
output embedding depends on all input embeddings. This all-to-all dependency is
what the authors mean when describing the model as ``bidirectional,'' which
departs from the more traditional left-to-right, right-to-left LSTM model.

When training BERT, input text is tokenized by the WordPiece
algorithm~\cite{wu2016google}, and two different types of training examples are
input. In the first, some tokens are randomly replaced with a masked reserve
token. The objective of the model during the unsupervised pre-training phase is
to predict the original token, using the rest of the input. In the second, two
sentences are supplied and, using the output embedding of the ``start-of-input''
character, the model must determine whether the second sentence followed the
first in the training data.

\noindent{\bf GPT}~\cite{radford2018improving} and {\bf
GPT-2}~\cite{radford2019language} both use a transformer-decoder stack of
\textit{masked} multi-headed self-attention. The mask, in this case, enforces
that the output embedding of token $i$ may only depend on inputs
$1,\dots,i)$. This masking formulation, which we adopt in this work,
restricts the GPT-models to function as pure language models. These models are
pre-trained through a generative objective. For each input sequence
$1,\cdots,n$, the model is input $1,\cdots,(n-1)$ and required to generate the
sequence $2,\cdots,n$. Due to the masked-self-attention layers, this means that
each prefix sequence of the input is simultaneously predicted each follow-up
word.

The major difference between the GPT and GPT-2 models is the larger training
corpus, which leads to state-of-the-art text
generation. In~\cite{radford2019language}, this model is even shown to improve
the state-of-the-art of other objectives such as question answering and
translation, even without a fine-tuning phase. Follow-up
work~\cite{petroni2019language} identifies that high-performance language models
like GPT-2 can even replace specialty knowledge-bases.
\fi

\noindent{\bf SciBert}~\cite{beltagy2019scibert} achieves state-of-the-art
performance across a range of scientific NLP benchmarks by retraining the
WordPeice tokenizer~\cite{wu2016google}, and a BERT model~\cite{devlin2019bert}
on 1.14-million papers collected by semantic scholar. Beltagy et al. demonstrate
that by performing unsupervised pre-training on this scientific dataset, they
are able to improve performance over the standard BERT-pre-trained weights on
their ultimate fine-tuned models for entity recognition, PICO extraction, text
classification, relation classification, and dependency parsing. These finding
make the case that scientific text is sufficiently dissimilar from that found in
general language to require custom models.

\noindent{\bf BioBert}~\cite{lee2019biobert} follows the same pattern as
SciBert, but pre-trains on the biomedical texts supplied by MEDLINE and
PubMedCentral. As opposed to SciBert, this method does not replace the
general-language training data supplied by English Wikipedia and BooksCorpus,
and instead appends both biomedical text databases. Lee et al.  explore the
resulting fine-tuned performance across a large range of small biomedical NLP
tasks, and find mixed results.  We interpret these results to indicate the
importance of finding training data that is not only sufficiently large, but
also relevant to the task at hand.

\noindent{\bf Wang et al.}~\cite{wang2019bert} explore the capacity for a BERT
model to effectively function as a Markov random field language model. This
technique takes advantage of the masked pre-training used in the base BERT model
to predict unknown tokens. This approach also departs from the traditional
language model described here as every sequence element determines the
probability of every other element.  Generation is performed by iterative
freezing highest-probability elements from within a fixed-length sequence of
initially free variables.

\noindent{\bf Ctrl}~\cite{keskar2019ctrl} is a conditional language generation
method that extends GPT by including ``control codes'' that prefix the sequence
of text elements.  For instance, each website represented in the training data
is represented by a code, and as a result generated text can switch styles based
on these prefixes. Additionally, various model functions, such as question
answering, are learned via generation with various codes. As a result, prefixing
questions with the respective code results in a higher probability assigned to
relevant answers. Furthermore, this work includes some multi-code prefixes, such
as ``Rating 5.0'' or ``Sentence Title'' to further condition the generated
result.  While the CTRL model is the most similar to the method presented here,
it has some key differences. Firstly, the CTRL model uses prefix tokens to
condition generated text, while we apply a shallow transformer-encoder stack. As
a result, the CTRL approach is limited in that training requires a strict set of
codes, or a small set of enumerable code-pairs. In contrast, the \sysname{}
approach allows the method proposed here to accept arbitrary-length sequences of
keywords as a condition.

\section{Future Challenges and Ethical Considerations}
\label{sec:future_work}

Many readers have likely heard of ``hoax'' paper generators similar to
Scigen~\cite{stribling2005scigen}. This particular project generates computer
science full-text articles by randomly sampling from a context-free grammar, and
has produced publications actually accepted by some venues.  This 15-year-old
system, however, is incapable of fooling but the least-observant of gate
keepers. 
However, high quality text generation introduces NLP to a range of challenges currently
posed by ``deepfake'' images.  These problematic pictures permeate the zeitgeist
and stir a response reaching further than computer
science~\cite{dolhansky2019deepfake}, extending into law~\cite{blitz2018lies},
culture~\cite{oxforddeepfakes}, and philosophy~\cite{floridi2018artificial}.
Meanwhile, misinformation spread by human actors online already cascades
throughout social network echo chambers at an alarming
rate~\cite{del2016spreading}. One needs very little imagination to conceive of
ways that the automatic generation of ``pseudo-science'' online could lead to
public distrust of the scientific community.

OpenAI is forming partnerships between computer-science and the social sciences
in order to understand these implications in
society~\cite{leibowicz2019appropriate}. One major challenge they note is a
distinct lack of ``correctness'' measures for text generation.  In completing
this work, we find that some correctness measures do exist, such as the SPICE
metric to judge image caption correctness~\cite{anderson2016spice}.
Unfortunately, this technique does not scale well to large knowledge bases as it
requires the graph of predicate arguments induced by reference sentences. Not
only are there a lack of methods to extract arguments from text, but we need to
find new algorithms for quantifying correctness for large graphs induced by all
of biomedical science.

Despite the potential for abuse, we designed \sysname{} with our own vision
toward enabling human-understandable hypothesis generation systems. For
instance, our model architecture could be conditioned on more generalized forms
of existing biomedical knowledge, such as semantic graph embeddings, in order to
produce textual descriptions of plausible future research directions. These
explanations could potentially persuade domain scientists to pursue new research
directions, as similar systems have already
done~\cite{aksenova2019inhibition,bakkar2018artificial}. However, these systems
require specialized analysis and introduce new cognitive burdens for scientists
to understand and act on their outputs.  If similar hypothesis generation
systems instead could produce human-readable arguments, then we could better
utilize the wealth of publicly available information, improve the productivity
of biomedical researchers, and ultimately find new treatments and cures for
people worldwide.

\section{Conclusions}
\label{sec:conclusion}

We present the Conditioned Biomedical Abstract Generation (\sysname{}) model for understanding scientific abstracts. We train this
model using publicly available biomedical data provide through MEDLINE to
predict text that is conditioned on publication year and arbitrary sets of author-supplied
keywords. This model leverages the transformer
architecture~\cite{vaswani2017attention}, featuring
a shallow condition
encoder, as well as a deep language model decoder. Using \sysname{}, and a range
of natural language generation metrics~\cite{sharma2017nlgeval}, we demonstrate
the need for such a domain-specialized model, as opposed to a larger more
general model like GPT-2.

We anticipate that conditioned language generation can be used to build new
applications in the biomedical domain, such as a hypothesis generation system
that produces textual descriptions of proposed new research directions. To do
so, the conditional aspect of the \sysname{} model will likely be a necessity.
However, we also acknowledge the ethical considerations behind the proliferation
of convincing scientific language generation models.
We provide the pre-trained model, over 13,000 generated abstracts, and all necessary
training and evaluation code to aid in exploration and reproduceability.

\bibliographystyle{plain}
\bibliography{main}

\end{document}